\documentclass[letterpaper]{article} 
\usepackage{aaai23}  
\usepackage{times}  
\usepackage{helvet}  
\usepackage{courier}  
\usepackage[hyphens]{url}  
\usepackage{graphicx} 
\usepackage{natbib}  
\usepackage{caption} 
\usepackage[linesnumbered, ruled]{algorithm2e}
\usepackage{amsmath}
\usepackage{amssymb}
\usepackage{subcaption}
\usepackage{cleveref}
\usepackage{subcaption,multirow,array,dsfont,ctable,xcolor}

\usepackage{newfloat}
\usepackage{listings}
\DeclareCaptionStyle{ruled}{labelfont=normalfont,labelsep=colon,strut=off} 
\title{Cyclically Disentangled Feature Translation for Face Anti-spoofing}
\author{
    Haixiao Yue\equalcontrib,
    Keyao Wang\equalcontrib,
    Guosheng Zhang\equalcontrib,
    Haocheng Feng,
    Junyu Han, \\
    Errui Ding,
    Jingdong Wang
}
\affiliations{
    Department of Computer Vision Technology(VIS), Baidu Inc.\\

    \{yuehaixiao, wangkeyao, zhangguosheng, fenghaocheng, hanjunyu, dingerrui\}@baidu.com, wangjingdong@outlook.com
}

\usepackage{bibentry}

\begin{document}

\maketitle

\begin{abstract}
Current domain adaptation methods for face anti-spoofing leverage labeled source domain data and unlabeled target domain data to obtain a promising generalizable decision boundary. However, it is usually difficult for these methods to achieve a perfect domain-invariant liveness feature disentanglement, which may degrade the final classification performance by domain differences in illumination, face category, spoof type, etc. In this work, we tackle cross-scenario face anti-spoofing by proposing a novel domain adaptation method called cyclically disentangled feature translation network (CDFTN). Specifically, CDFTN generates pseudo-labeled samples that possess: 1) source domain-invariant liveness features and 2) target domain-specific content features, which are disentangled through domain adversarial training. A robust classifier is trained based on the synthetic pseudo-labeled images under the supervision of source domain labels. We further extend CDFTN for multi-target domain adaptation by leveraging data from more unlabeled target domains. Extensive experiments on several public datasets demonstrate that our proposed approach significantly outperforms the state of the art. 
Code and models are available at https://github.com/vis-face/CDFTN.
\end{abstract}

\begin{figure}[!t]
    \centering
    \includegraphics[width=8.2cm]{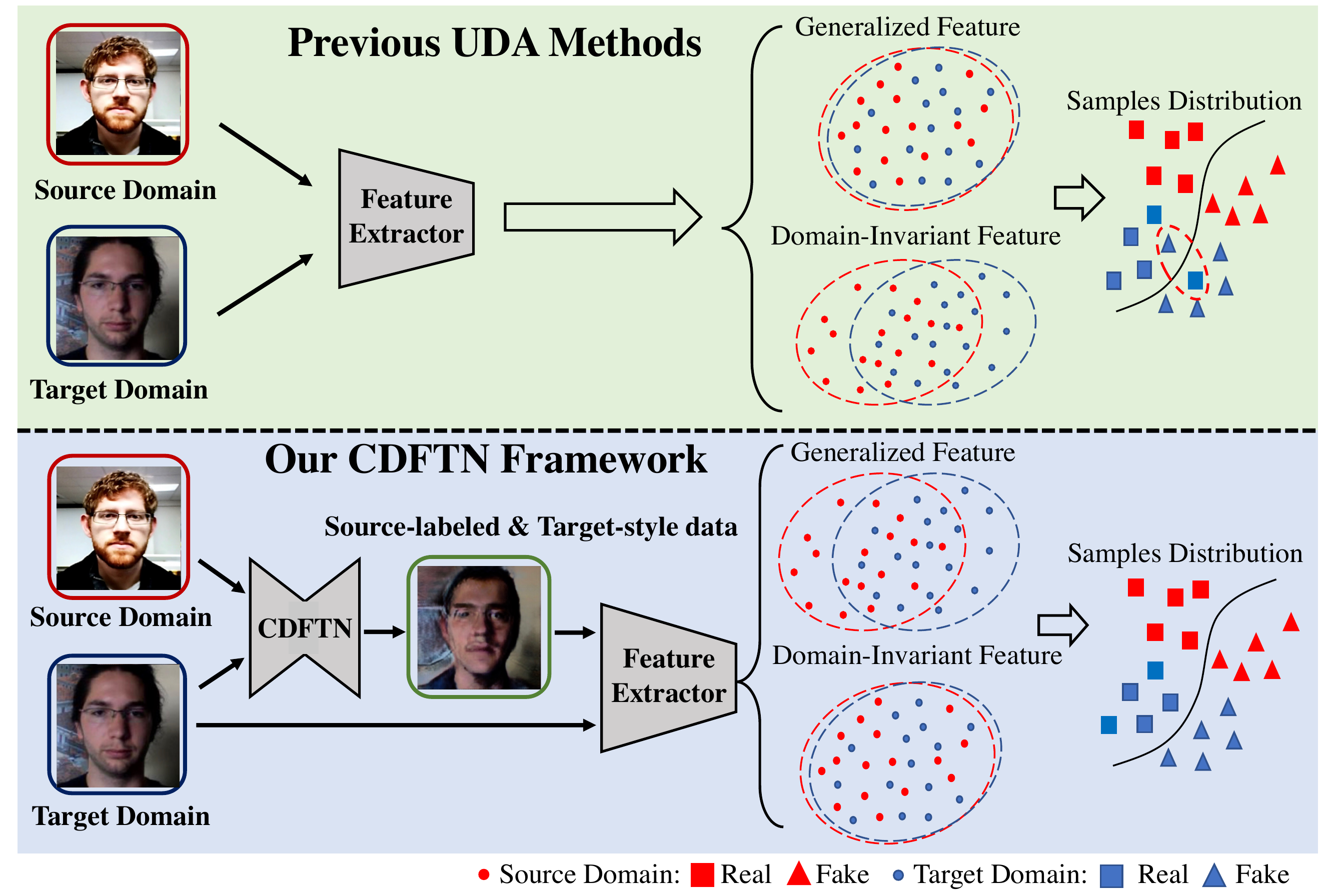}
    \caption{Framework comparison among our proposed method and previous UDA methods. The previous UDA methods translate generalized features which do not fit classification well. Our CDFTN method can disentangle and extract domain-invariant liveness features for better classification on the target domain.}
    \label{fig:compare}
\end{figure}

\section{Introduction}
In recent years, face recognition has become a prominent technique for identity authentication and has been widely used in our lives. However, existing face recognition systems are vulnerable to face presentation attacks such as printed photos (i.e. print attack), digital images or videos (i.e. replay attack), and 3D facial masks (i.e. 3D mask attack), etc. In light of this urgency, the development of face anti-spoofing (FAS) technique has shown increasing significance in face recognition systems.

Various approaches have been proposed to train a generalizable classifier on different FAS scenarios. Most of existing solutions exploit multi-source domain generalization (DG) methods to learn a domain agnostic representation that aligns discriminative features on source domains, such that the model can generalize to the target domain without accessing its data. In practice, however, labeling the multiple source domains is time-consuming and laborious. Moreover, since we have access to plenty of unlabeled facial image data from an existing face recognition system, domain adaptation (DA) forms a natural learning framework for FAS. DA methods seek to aid cross-scenario FAS by extracting domain-indistinguishable feature representations from both labeled source data and unlabeled target data. Therefore, they can exploit rich information in the unlabeled target domain and obtain a more robust decision boundary.

On the FAS task, domain-invariant liveness feature extraction and translation are crucial to the final classification performance. Based on this thought, we assume that facial images in the FAS task could be mapped into two latent feature spaces: 1) a domain-invariant liveness feature space that represents the live-or-spoof attribute and 2) a domain-specific content feature space that represents face category, background environment, camera, and illumination, which is irrelevant to live-or-spoof classification. In this work, we propose a simple but effective cyclically disentangled feature translation network (CDFTN) to deal with the cross-scenario face anti-spoofing problem. Figure \ref{fig:compare} presents how to learn CDFTN with source and target domains.

In specific, CDFTN aims at generating cross-domain pseudo-labeled images, which is achieved by swapping the domain-invariant liveness features and the domain-speciﬁc content features from different domains. The labels of synthetic images are assigned to be the same as those of source domain images. To obtain disentangled representations along with effective generators, we employ GAN-like \cite{NIPS2014_5ca3e9b1} discriminators to conduct domain adversarial training. In addition, cyclic reconstruction and latent reconstruction are used to guarantee the effectiveness of disentangled feature translation. We ﬁnally train a robust classifier on the generated pseudo-labeled images and evaluate the trained classifier directly on the target dataset for testing. In contrast to existing DA-based FAS methods \cite{li2018unsupervised,wang2021unsupervised} that directly make decisions based on the exacted domain-invariant features, we instead use these features to synthesize training samples and obtain a discriminative classifier on synthetic pseudo-labeled training images. 

Given the practical scenario that unlabeled target datasets from multiple domains are accessible, we extend our proposed method from single-source $\rightarrow$ single-target feature translation to single-source $\rightarrow$ multiple-targets translation. This extended network allows domain-invariant liveness features to transfer within multiple domains and generates pseudo-labeled images for each target domain. We finally train a robust classifier based on all pseudo-labeled images.  

The contributions are summarized as follows:
\begin{itemize}
    \item To tackle the cross-scenario FAS problem based on labeled source domain data and unlabeled target domain data, we propose to generate pseudo-labeled images to train a generalizable classifier.
    \item We design a novel feature translation framework using disentangled representation learning based on domain adversarial training; we also extend the framework from single-source $\rightarrow$ single-target to single-source $\rightarrow$ multiple-targets feature translation.
    \item Given unlabeled target domain data without other depth or temporal information, our approach achieves superior performance over the state-of-the-art methods.
\end{itemize}

\begin{figure*}
    \centering
    \includegraphics[width=0.98\textwidth]{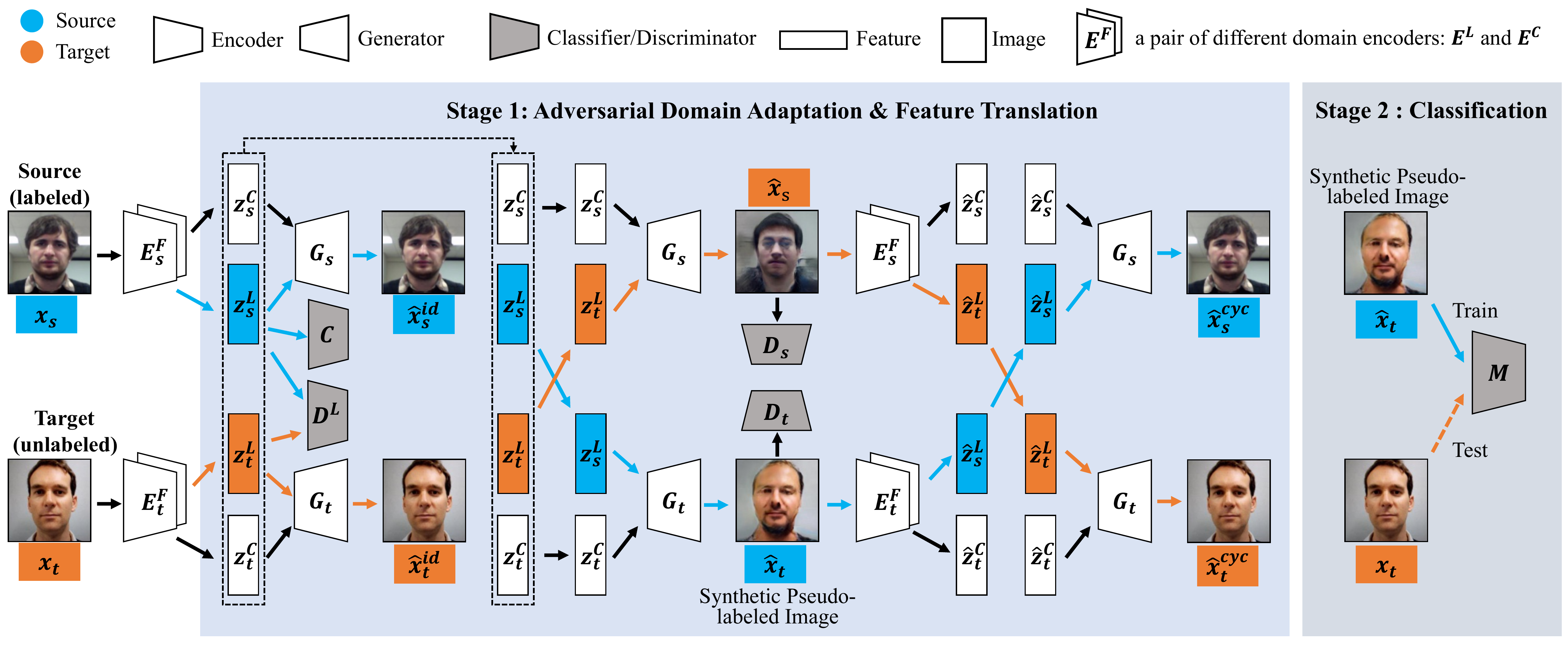}
    \caption{An overview of the model architecture. The network in Stage 1 visualizes the following functions of our model: 1) by applying a pair of different encoders (i.e. $E^L$ and $E^C$) in each domain, our model factorizes the liveness features and content features. 2) by using liveness discriminator $D^L$ and classifier $C$, our model extracts domain-invariant liveness features through adversarial domain adaptation. 3) by exchanging $z^L$ between domains, the generator $G$ generates images based on cross-domain feature representations. Generated images in target domain $\hat{x}_t$ would be passed to Stage 2 to obtain a domain-adapted classifier $M$.}
    \label{fig:architecture}
\end{figure*}

\section{Related Works}
\subsection{Face Anti-spoofing Methods}
Traditional face anti-spoofing methods extract hand-crafted features such as LBP \cite{maatta2011face}, HOG \cite{komulainen2013context}, and SIFT \cite{patel2016secure} to capture spoof patterns. With the recent development of deep learning, researchers use convolutional neural networks (CNN) to exploit deeper discriminative feature representations for face presentation attack detection \cite{atoum2017face,feng2020learning,liu2018learning,yu2020face,yu2020searching}. CNN is used as a feature extractor for presentation attack detection, which is fine-tuned from ImageNet-pretrained ResNet \cite{he2016deep} and VGG \cite{simonyan2014very}. \cite{feng2020learning} proposes a residual-learning framework to learn the discriminative spoof cues in the framework of anomaly detection. \cite{yu2020face} regards face anti-spoofing problem as a material recognition problem. \cite{yu2020searching} proposes a framework based on central difference convolution. Several recent works \cite{feng2016integration,wang2020deep,xu2015learning} utilize LSTM or GRU to extract discriminative features in a stochastic manner for better classification performance. \cite{CelebA-Spoof} trains the network in a unified multi-task framework. Although FAS has achieved great development in the past decade, the previously discussed methods only focus on a single domain and cannot generalize well on others.

\subsection{Unsupervised Domain Adaptation}
Several recent domain generalization methods \cite{jia2020single,quan2021progressive,shao2019multi,wang2020cross,Liu_2022_AAAI,wang2022domain} have achieved remarkable development on cross-domain FAS problem. Nevertheless, amount of unlabeled target data are available in some scenarios. But the data labeling work is laborious and time-consuming. Unsupervised domain adaptation (UDA) provides an alternative way by transferring knowledge from source domain to target domain. There are various types of proposed methods: (i) minimizing discrepancy between source and target domains sub-spaces \cite{gong2012geodesic,long2013transfer,pan2010domain,sun2015subspace}; (ii) domain confusion with adversarial approach \cite{donahue2016adversarial,kim2017learning,tzeng2017adversarial}; (iii) domain transformation with image-to-image translation \cite{lee2018diverse,yi2017dualgan,zhu2017unpaired}. However, there are only a few existing works focusing on FAS in the UDA framework, within which \cite{li2018unsupervised} proposed a UDA framework for presentation-attack detection by minimizing Maximum Mean Discrepancy (MMD) \cite{gretton2012kernel}; \cite{jia2021unified,wang2021unsupervised} proposed a UDA framework with adversarial training to improve the generalization capability of FAS model on target scenarios. \cite{wang2021self} proposed a self-domain adaptation framework to leverage the unlabeld test domain data at inference. However, it is usually difficult for these methods to achieve a perfect domain-invariant liveness feature disentanglement, which may degenerate the final classification performance.

\subsection{Disentangled Representation Learning}
Deep neural networks are known to extract features where multiple hidden features are highly entangled \cite{peng2020domain2vec,zhuang2015supervised}. Disentangled representation learning focuses on extracting relevant features given a large variation of each dataset as well as irrelevant features that depict the relations between different domains. Recently, more and more state-of-the-art methods utilize generative adversarial networks (GAN) \cite{NIPS2014_5ca3e9b1} and variational autoencoders (VAE) \cite{kingma2013auto} to learn disentangled representations given their success on image generation tasks. In the scenario of FAS, \cite{liu2020disentangling} designs a novel adversarial learning framework to disentangle the spoof traces from input faces as a hierarchical combination of patterns at multiple scales. \cite{wang2020cross} proposed a disentangled representation learning for cross-domain face presentation attack detection. \cite{zhang2020face} proposed a CNN architecture with the process of disentanglement and a combination of low-level and high-level supervision to improve generalization capabilities. Despite great achievement in FAS based on multi-level supervision, those methods cannot disentangle domain-specific and domain-invariant presentation attack representations thus would still overfit on training set.

\section{Proposed Method}
The full network is illustrated in Figure \ref{fig:architecture}. In this section, we introduce our problem statement and objectives, and then go through our proposed method in more details.

Suppose we have source dataset with $n_s$ labeled examples $\mathcal{S}=\{(x^i_s, y^i_s)\}^{n_s}_{i=1} \in X_s$, where $x^i_s \in \mathbb{R}^{H\times W \times 3}$ denotes the $i^{th}$ source image and $H, W$ stand for the height and width of $x_s^i$. Given the scenario of FAS, $y^i_s$ is a one-hot vector with two elements corresponding to its label representing genuine or spoof. Similarly, we have a target dataset which includes $n_t$ unlabeled examples $\mathcal{T}=\{(x^i_t)\}^{n_t}_{i=1} \in X_t$ where $x^i_t \in \mathbb{R}^{H\times W \times 3}$ denotes the $i^{th}$ target image with the same size as $x^i_s$.
Due to domain shift, the marginal distributions of source and target datasets are different, i.e., $P_{\mathcal{S}}(X_s) \neq P_{\mathcal{T}}(X_t)$.
The ultimate goal of our proposed method is to train a classifier that could effectively estimate $P_{\mathcal{T}}(Y_t|X_t)$ without the prior knowledge of $P_\mathcal{T}(Y_t)$.

\subsection{Cross-Domain Feature Disentanglement} \label{sec:uda}
We first learn the common liveness-related features from source and target domains. Figure \ref{fig:architecture} presents our network. Given the scenario of face anti-spoofing, we assume that both domains share common liveness properties despite the apparent differences between domains. We aim at mapping inputs from both domains into a common space.
We first apply a pair of liveness encoder $E_s^L$ and $E_t^L$ to extract domain-invariant liveness features $z^L_s$ and $z^L_t$, and a pair of content encoder $E_s^C$ and $E_t^C$ to extract domain-specific content features $z^C_s$ and $z^C_t$. To determine the domain affiliation of extracted features $z_s^L$ and $z_t^L$, we conduct domain adversarial training by applying a GAN-like \cite{NIPS2014_5ca3e9b1} discriminator $D^L$. We formulate the domain adversarial loss as follows:

\begin{equation} \label{eq:lganloss} 
\begin{split}
    \min_{E_s^L,E_t^L}\max_{D^L}\mathcal{L}_{D^L}= &\mathbb{E}_{x_s \sim P_{\mathcal{S}}(x_s)}[\log D^L(E_s^L(x_s))]\\ +  &\mathbb{E}_{x_t \sim P_{\mathcal{T}}(x_t)}[\log (1-D^L(E_t^L(x_t)))].
\end{split}
\end{equation}

The discriminative property of encoded feature $z_s^L$ is determined by labels of source domain. The cross-entropy loss of liveness feature from domain $\mathcal{S}$ is formulated as follows:

\begin{equation} \label{eq:clsloss}
    \min_{E_s^L,C}\mathcal{L}_{cls^L}=
    -\mathbb{E}_{(x_s,y_s) \sim P_{\mathcal{S}}(x_s,y_s)} [y_s\log C(E_s^L(x_s))],
\end{equation}
where $C$ is a binary classifier. 
A pair of decoders $G_s$ and $G_t$ is applied to reconstruct the extracted features back to original input image at the pixel level. We formulate reconstruction loss $\mathcal{L}^{re}$ as:

\begin{equation} \label{eq:idloss}
\begin{split}
    \min_{G_d,E_d^f}\mathcal{L}^{re}=&\mathbb{E}_{x_s \sim P_{\mathcal{S}}(x_s)}[|| G_s(E_s^L(x_s),E_s^C(x_s))-x_s||_{1}]\\
    +&\mathbb{E}_{x_t \sim P_{\mathcal{T}}(x_t)}[||G_t(E_t^L(x_t),E_t^C(x_t))-x_t||_{1}],
\end{split}
\end{equation}

\noindent where $d \in \{s,t\}$ and $f \in \{L,C\}$.

\subsection{Single-target Feature Translation} \label{sec:uft}
Besides learning a common space for domain-invariant features, we also transfer them from labeled source domain to unlabeled target domain and use them to generate pseudo-labeled images. The architecture of feature translation framework is shown in the Figure \ref{fig:architecture}. We further train a robust classifier on pseudo-labeled images and evaluate the classifier directly on original images in target domain.
To generate ideal pseudo-labeled images, the extracted domain-invariant liveness features are swapped between domains and concatenated with corresponding domain-specific content features; then the concatenated feature is fed into $G_s$ and $G_t$ to construct fake images: $\hat{x}_t=G_t(z_s^L,z_t^C)$, $\hat{x}_s=G_s(z_t^L,z_s^C)$ and $\hat{x}_t$ denotes pseudo-labeled images.
To synthesize authenticate images, we further add a pair of discriminators $D_s$ and $D_t$ to differentiate between $\hat{x}_{s}$ and $x_{s}$, $\hat{x}_{t}$ and $x_{t}$. The adversarial loss is formulated as:

\begin{equation} \label{eq:dganloss} 
\begin{split}
    \max_{D_s,D_t}\mathcal{L}^{adv}_{D}=&\mathbb{E}_{x_s \sim P_{\mathcal{S}}(x_s)}[\log D_s(x_s)+\log (1-D_s(\hat{x}_s))] \\
    +&\mathbb{E}_{x_t \sim P_{\mathcal{T}}(x_t)}[\log D_t(x_t)+\log (1-D_t(\hat{x}_t))].
\end{split}
\end{equation}

\noindent Inspired by \cite{zhu2017unpaired}, the trained feature encoder and decoder functions should be able to bring $x$ back to the original input, which is called \emph{cycle consistency}. This statement holds an intuition that if the mapping functions are able to transfer features from domain $\mathcal{S}$ to domain $\mathcal{T}$, then they are also expected to bring the same features back to the original domain. The cycle-consistency loss is formulated as:

\begin{equation} \label{eq:cycloss}
\begin{split}
    \min_{G_{d},E_{d}^f}\mathcal{L}^{cyc}=&\mathbb{E}_{x_s \sim P_{\mathcal{S}}(x_s)}[|| G_s(E_t^L(\hat{x}_t),E_s^C(\hat{x}_s))-x_s||_{1}]\\
    +&\mathbb{E}_{x_t \sim P_{\mathcal{T}}(x_t)}[||G_t(E_s^L(\hat{x}_s),E_t^C(\hat{x}_t))-x_t||_{1}].
\end{split}
\end{equation}

\noindent Finally, to enforce the liveness features extracted from $x_s$($x_t$) and $\hat{x}_t$($\hat{x}_s$) to be unchanged after translation, we also apply the reconstruction loss between the domain-invariant liveness features as follows:

\begin{equation} \label{eq:latloss}
\begin{split}
    \min_{E_s^L,E_t^L}\mathcal{L}^{lat}=&\mathbb{E}_{x_s \sim P_{\mathcal{S}}(x_s)}[||E_s^L(\hat{x}_s)-E_t^L(x_t)||_{1}]\\
    +&\mathbb{E}_{x_t \sim P_{\mathcal{T}}(x_t)}[||E_t^L(\hat{x}_t)-E_s^L(x_s)||_{1}].
\end{split}
\end{equation}

\subsection{Full Objective Function} 
The whole network is trained step by step and we denote this training algorithm as \textbf{S}ingle \textbf{S}ource to \textbf{S}ingle \textbf{T}arget which abbreviated as \textbf{SS2ST}. Optimization of proposed loss functions could be summarized as the following two stages:

\noindent \textbf{Stage 1} is shown in the first columns of Figure \ref{fig:architecture}. We first simultaneously optimize Equations (\ref{eq:lganloss})-(\ref{eq:latloss}) to achieve both liveness feature domain adaptation and inter-domain translation. The objective function is:

\begin{equation} \label{eq:stage1loss}
\begin{split}
    &\min_{G_{d},E_{d}^f}\max_{D^L,D_{d}}\mathcal{L}^{\text{stage1}}\\
    &=\mathcal{L}_{cls^L} + \lambda_1 \mathcal{L}_{D^L} + \lambda_2 \mathcal{L}_D^{adv} + \lambda_3 \mathcal{L}^{re} + \lambda_4 \mathcal{L}^{cyc} + \lambda_5 \mathcal{L}^{lat},
\end{split}
\end{equation}
\noindent where $\big \{\lambda_k\big\}_{k=1}^5$ are hyper-parameters. 

\noindent \textbf{Stage 2} is the training step as shown in the second columns of Figure \ref{fig:architecture}. After Equation (\ref{eq:stage1loss}) converges, all parameters in CDFTN are fixed to generate pseudo-labeled images. The ultimate classifier (denoted as $M$) is trained on generated image $\hat{x}_t$. 
We adopt two different architectures of the classifier for comparisons. For the first one, we select the LGSC \cite{feng2020learning} as the classifier since the method presents the best performance. The second one is the binary classifier of ResNet-18 \cite{he2016deep}. We denote these two different classifiers by L and R for short
in the following (i.e., CDFTN-L and CDFTN-R).  According to \cite{feng2020learning}, the proposed loss consists of a auxiliary binary classification loss $\mathcal{L}_a$, a spoof cue $L_1$ regression loss $\mathcal{L}_r$, and triplet loss $\mathcal{L}_{tri}$, while only binary classification loss $\mathcal{L}_a$ is considered for ResNet-18\cite{he2016deep}. Therefore, the loss of stage 2 is defined as:

\begin{equation} \label{eq:stage2loss}
    \min \mathcal{L}^{\text{stage2}}=\alpha_1 \mathcal{L}_a+\alpha_2 \mathcal{L}_r+\alpha_3 \mathcal{L}_{tri},
\end{equation}

\noindent where $\big \{\alpha_k\big\}_{k=1}^3$ are hyper-parameters defined in \cite{feng2020learning} to control the relative importance of each objectives.

\begin{figure}
    \centering
    \includegraphics[height=5cm]{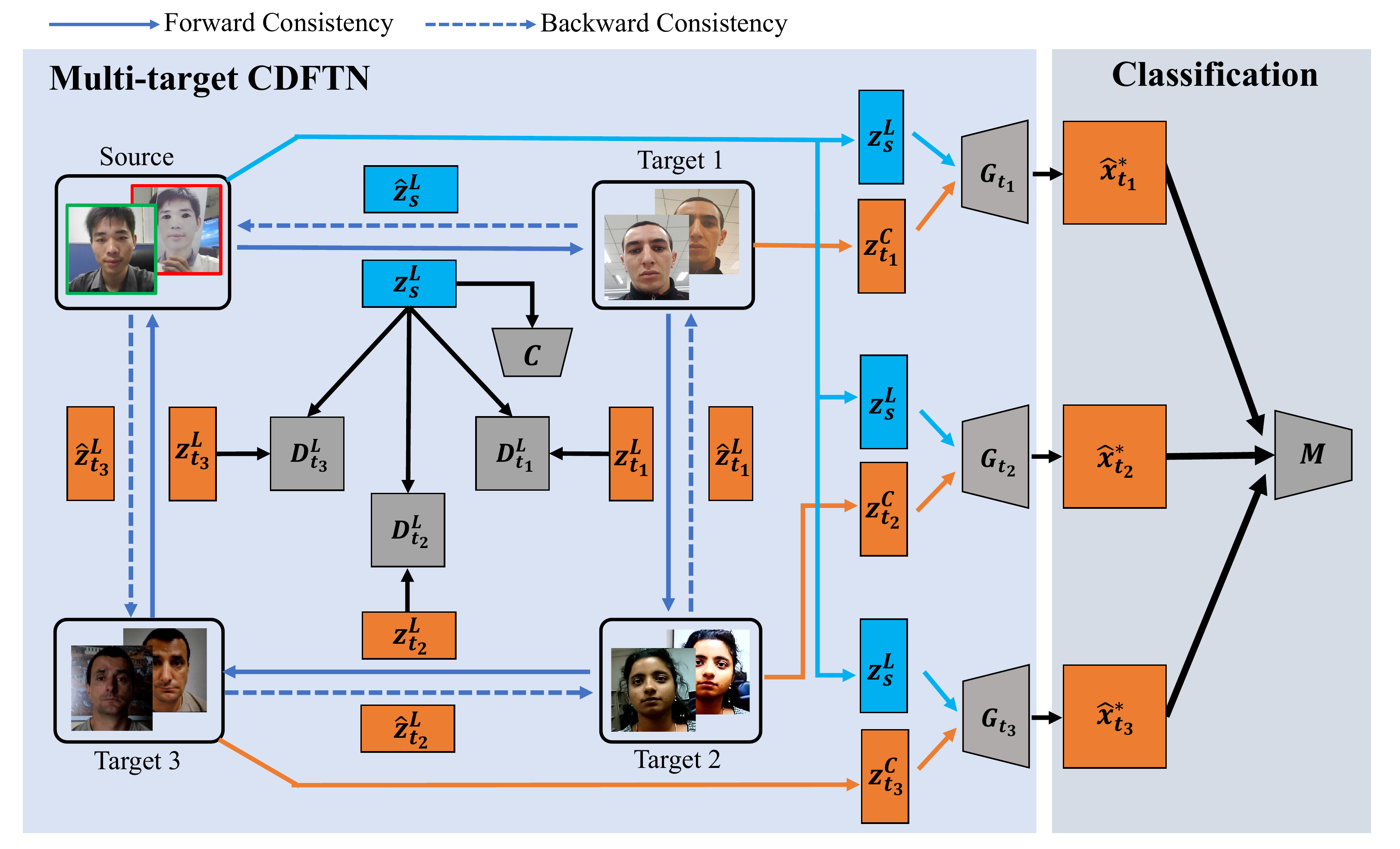}
    \caption{Network structure of multi-target translation where the model includes one labeled source domain and three unlabeled target domains corresponding to experimental settings. Forward consistency visualizes the path that $z^L$ features are transferred through whereas backward consistency presents the path that $z^L$ transferred back to the original domain. The synthetic pseudo-labeled images from multiple target domains ($\hat{x}_{t_1}$, $\hat{x}_{t_2}$, $\hat{x}_{t_3}$) are combined to train a classifier $M$.}
    \label{fig:multi_arch}
\end{figure}

\SetKwInOut{Notation}{Notation}
\begin{algorithm}[!h]
 \KwIn{Source $\mathcal{S}$; Target $\{\mathcal{T}_i\}_{i=1}^N$; Feature encoder $E_s^L$, $\{E_{t_i}^L\}_{i=1}^N$, $E_s^C$, $\{E_{t_i}^C\}_{i=1}^N$; Decoder $G_s$, $\{G_{t_i}\}_{i=1}^N$; Discriminator $\{D^L_{t_i}\}_{i=1}^N$, $D_s$, $\{D_{t_i}\}_{i=1}^N$; Classifier $C$, $M$}
\KwOut{well-trained $M^{*}$}
\Notation{$\mathcal{D}=\{s,\{t_i\}_{i=1}^N\}$, $\mathcal{F}=\{L,C\}$}
 \While{stage 1 not converged}{
  Random sample a mini-batch $X_s$ from $\mathcal{S}$ and mini-batches $X_{t_1},...,X_{t_N}$ from $\mathcal{T}_1,...,\mathcal{T}_N$\;
  \textbf{\emph{Adversarial domain adaptation}}\;
  $z_s^L=E_s^L(X_s)$, $\{z_{t_i}^L=E_{t_i}^L(X_{t_i})\}_{i=1}^N$\;
  Update $C$, $E_s^L$ with (\ref{eq:clsloss})\;
  $\forall d \in \mathcal{D}$, update $E_d^L$, $\{D^L_{t_i}\}_{i=1}^N$ with (\ref{eq:lganloss})\;
  \textbf{\emph{Image generation}}\;
  $\hat{x}_s=G_s(z_{t_N}^L,z_s^C)$\;
  $\hat{x}_{t_1}=G_{t_1}(z_s^L,z_{t_1}^C)$, $\{\hat{x}_{t_i}=G_{t_i}(z_{t_{i-1}}^L,z_{t_i}^C)\}_{i=2}^N$\;
  $\forall d \in \mathcal{D}$, update $E_d^C$, $G_d$ with (\ref{eq:dganloss}, \ref{eq:idloss}), $D_d$ with (\ref{eq:dganloss})\;
  \textbf{\emph{Cycle Consistency}}\;
  $\hat{x}^{cyc}_s=G_s(E_{t_1}^L(\hat{x}_{t_1}),E_s^C(\hat{x}_s))$\; $\{\hat{x}^{cyc}_{t_i}=G_{t_i}(E_{t_{i+1}}^L(\hat{x}_{t_{i+1}}),E_{t_i}^C(\hat{x}_{t_i}))\}_{i=1}^{N-1}$\;
  $\hat{x}^{cyc}_{t_N}=G_{t_N}(E_s^L(\hat{x}_s),E_{t_N}^C(\hat{x}_{t_N}))$\;
  $\forall d \in \mathcal{D}$, $\forall f \in \mathcal{F}$, update $E_d^f$, $G_d$ with (\ref{eq:cycloss}, \ref{eq:latloss})
 }
 \While{stage 2 not converged}{
 $\forall d \in \mathcal{D}$, $\forall f \in \mathcal{F}$, fix $E_d^f$, $G_d$, $D_d$ and $\{D^L_{t_1}\}_{i=1}^N$\;
 \textbf{\emph{Pseudo-labeled image synthesis}}\;
 $\{\hat{x}_{t_i}^{*}=G_{t_i}(z_s^L,z_{t_i}^C)\}_{i=1}^N$\;
 \textbf{\emph{Classification}}\;
 Update $M$ with (\ref{eq:stage2loss})
 }
 \caption{Training Procedure of CDFTN}
 \label{alg}
\end{algorithm}

\subsection{Multi-target Feature Translation} \label{sec:expansion}
 We propose to extend CDFTN to include multiple unlabeled target datasets. The network is presented in Figure \ref{fig:multi_arch}. Suppose we have one labeled source domain and $N$ unlabeled target domains, we use different liveness encoders $E_s^L$ and $\{E^L_{t_i}\}_{i=1}^N$ to extract liveness features $z_s^L$ and $\{z^L_{t_i}\}_{i=1}^N$ and apply $N$ separate liveness discriminators $\{D^L_{t_i}\}_{i=1}^N$ for domain adversarial training.
 
 The extracted liveness features $z^L$ are transferred via a forward loop from source domain $\mathcal{S}$ to the target domain $\mathcal{T}_1$, from $\mathcal{T}_1$ to target domain $\mathcal{T}_2$, \ldots, and finally from target domain $\mathcal{T}_N$ back to source domain $\mathcal{S}$. We use different content encoders $E_s^C$ and $\{E^C_{t_i}\}_{i=1}^N$ to extract domain-specific content features $z_s^C$ and $\{z^C_{t_i}\}_{i=1}^N$ and apply different decoders $G_s$ and $\{G_{t_i}\}_{i=1}^N$ in each domain to generate images $\{\hat{x}_{t_i}\}_{i=1}^N$. Similar to the training step as Figure \ref{fig:architecture} mentioned, we synthesize pseudo-labeled images $\{\hat{x}_{t_i}^{*}\}_{i=1}^N$ and train a robust classifier $M$ on all pseudo-labeled images. Similar to SS2ST, we denote this model extension module as \textbf{S}ingle \textbf{S}ource to \textbf{M}ultiple \textbf{T}argets short for \textbf{SS2MT}. 
The training steps are shown in Algorithm \ref{alg}. When $N=1$,  \ref{alg} reduces to the SS2ST method.
 
\section{Experiments}

\subsection{Databases}
\label{sec:databases}
We provide our evaluations on four publicly available databases for cross-domain FAS: CASIA-MFSD \cite{zhang2012face} (C for short), Replay-Attack \cite{chingovska2012effectiveness} (I for short), MSU-MFSD \cite{wen2015face} (M for short) and Oulu-NPU \cite{boulkenafet2017oulu} (O for short). For SS2ST, we regard each dataset as a single domain to evaluate our model by selecting a source and target domain pair. 
The whole source domain dataset and the training set of target domain will be utilized in the training process.
Therefore, in SS2ST we will experiment upon the following 12 scenarios: C $\rightarrow$ I, C $\rightarrow$ M, C $\rightarrow$ O, I $\rightarrow$ C, I $\rightarrow$ M, I $\rightarrow$ O, M $\rightarrow$ C, M $\rightarrow$ I, M $\rightarrow$ O, O $\rightarrow$ C, O $\rightarrow$ I and O $\rightarrow$ M. For SS2MT, among datasets C, I, M and O, we also select one as labeled source and the other three as unlabeled targets. Thus in SS2MT we have 4 scenarios: C $\rightarrow$ I\&M\&O, I $\rightarrow$ C\&M\&O, M $\rightarrow$ C\&I\&O and O $\rightarrow$ C\&I\&M.

\subsection{Evaluation Metrics}
We use the Half Total Error Rate (HTER), which is the mean of False Rejection Rate (FRR) and False Acceptance Rate (FAR) for cross-domain FAS:
\begin{equation} \label{eq:hter}
    HTER = \frac{FRR+FAR}{2}.
\end{equation}
We also export Area Under the Curve (AUC) for quantitative comparison between SS2ST and SS2MT.

\begin{table*}[!t]
    \centering
    \renewcommand\tabcolsep{2.5pt}
    \resizebox{2.1\columnwidth}{22.2mm}{
    \begin{tabular}{cccccccccccccc}
    \specialrule{.1em}{.05em}{.05em}
    \noalign{\vskip 1.5pt}
    \textbf{Method} & 
    \textbf{C} $\rightarrow$ \textbf{I} & 
    \textbf{C} $\rightarrow$ \textbf{M} & 
    \textbf{C} $\rightarrow$ \textbf{O} &
    \textbf{I} $\rightarrow$ \textbf{C} & 
    \textbf{I} $\rightarrow$ \textbf{M} & 
    \textbf{I} $\rightarrow$ \textbf{O} &
    \textbf{M} $\rightarrow$ \textbf{C} & 
    \textbf{M} $\rightarrow$ \textbf{I} & 
    \textbf{M} $\rightarrow$ \textbf{O} &
    \textbf{O} $\rightarrow$ \textbf{C} & 
    \textbf{O} $\rightarrow$ \textbf{I} & 
    \textbf{O} $\rightarrow$ \textbf{M} &
    \textbf{Avg.}\\

    \noalign{\vskip 1.5pt}
    \hline
    \noalign{\vskip 1.5pt}
    ADDA \cite{tzeng2017adversarial} &41.8&36.6&-&49.8&35.1&-&39.0&35.2&-&-&-&-&39.6\\
    DRCN \cite{ghifary2016deep} &44.4&27.6&-&48.9&42.0&-&28.9&36.8&-&-&-&-&38.1\\
    DupGAN \cite{hu2018duplex} &42.4&33.4&-&46.5&36.2&-&27.1&35.4&-&-&-&-&36.8\\
    KSA \cite{li2018unsupervised}&39.3&15.1&-&12.3&33.3&-&9.1&34.9&-&-&-&-&24.0\\
    DR-UDA \cite{wang2021unsupervised}&15.6&9.0&28.7&34.2&29.0&38.5&16.8&3.0&30.2&19.5&25.4&27.4&23.1\\
    MDDR \cite{wang2020cross}&26.1&20.2&\textbf{24.7}&39.2&23.2&33.6&34.3&8.7&31.7&21.8&27.6&22.0&26.1\\
    ADA \cite{wang2019improving}&17.5&9.3&29.1&41.5&30.5&39.6&17.7&5.1&31.2&19.8&26.8&31.5&25.0\\
    USDAN-Un \cite{jia2021unified}&16.0&9.2&-&30.2&25.8&-&13.3&3.4&-&-&-&-&16.3\\
    \specialrule{.1em}{.05em}{.05em}
    \noalign{\vskip 1.5pt}
    \textbf{CDFTN-R} &5.4&14.4&32.5&\textbf{8.7}&12.9&\textbf{25.1}&13.5&5.6&28.2&\textbf{10.0}&\textbf{2.2}&7.1&13.8\\
    \textbf{CDFTN-L}&\textbf{1.7}&\textbf{8.1}&29.9&11.9&\textbf{9.6}&29.9&\textbf{8.8}&\textbf{1.3}&\textbf{25.6}&19.1&5.8&\textbf{6.3}&\textbf{13.2}\\
    \specialrule{.1em}{.05em}{.05em}
    \end{tabular}} \\
    \caption{Cross-database testing results in comparison on four testing domains between our method and other methods. CDFTN-R and CDFTN-L denote using ResNet-18 and LGSC as the binary classifier respectively.}
    \label{tab:result_da}
\end{table*}

\subsection{Implementation Details} \label{sec:implement}
In our experiment, we crop the human faces out of original images. For datasets that offer no face location ground truth, we use the Dlib \cite{kim2017learning} toolbox as the face detector. We resize our input image to $224\times224\times3$, where we extract the RGB channels of each image. Training examples are resampled to keep the live-spoof ratio to 1:1. We implement Equation (\ref{eq:stage1loss}) and Equation (\ref{eq:stage2loss}) separately. The Adam optimizer \cite{kingma2014adam} is applied for both stages. In Stage 1, the learning rate is set as $1 \times 10 ^{-3}$ and betas of optimizer are set to $(0.5, 0.999)$; we choose values of $\lambda_1$, $\lambda_2$, $\lambda_3$, $\lambda_4$ as $1$, $1$, $10$, $10$, respectively; the batch size is set to be 2 and the training process lasts for 30 epochs. 
In Stage 2, we set $\alpha_1$, $\alpha_2$, $\alpha_3$ to be the same as \cite{feng2020learning}. During training, the batch size is set to be 32 and the classifier would be trained for 5 epochs.

\begin{table}[!t]
    \centering
    \renewcommand\tabcolsep{5pt}
    \begin{tabular}{ccccc}
    \specialrule{.1em}{.05em}{.05em}
    \multicolumn{2}{c}{\textbf{Method}} & \textbf{SS2BT} & \textbf{SS2ST} & \textbf{SS2MT} \\ 
    \hline
    \multirow{2}{*}{\textbf{C$\rightarrow$I\&M\&O}} & HTER(\%) & 15.9  & 13.3  & \textbf{13.1}  \\
                           & AUC(\%)  & 89.5  & 90.8  & \textbf{92.1}  \\ 
    \multirow{2}{*}{\textbf{I$\rightarrow$C\&M\&O}} & HTER(\%) & 24.0  & 18.1  & \textbf{17.0}  \\
                           & AUC(\%)  & 83.0  & 89.1  & \textbf{90.2}  \\ 
    \multirow{2}{*}{\textbf{M$\rightarrow$C\&I\&O}} & HTER(\%) & 19.6  & 11.6  & \textbf{11.3}  \\
                           & AUC(\%)  & 88.2  & 93.2  & \textbf{93.5}  \\ 
    \multirow{2}{*}{\textbf{O$\rightarrow$C\&I\&M}} & HTER(\%) & 19.7  & 10.4  & \textbf{7.8}   \\
                           & AUC(\%)  & 86.6  & 95.6  & \textbf{97.4}  \\ 
    \specialrule{.1em}{.05em}{.05em}
    \end{tabular}
    \caption{Cross database testing results on the combination of three target domains. The results are reported as the average HTER and AUC scores. Specifically, the HTER scores of SS2ST are accordingly the arithmetic average of those shown in Table \ref{tab:result_da}.}
    \label{tab:result_compare_ours}
\end{table}

\subsection{Comparison with state-of-art-methods}

We compare our proposed method with other DA methods. The results are presented in Table \ref{tab:result_da}. It can be found that the result of our proposed framework outstrips other methods. This is because our method encourages feature translation instead of simply domain adaptation that leverages target domain-specific features to obtain a robust and generalizable classifier. However, our method does not significantly improve cross-database testing performance when the source domain is CASIA-MFSD and the target domain is Oulu-NPU. There two possible reasons for this result. (1) Oulu-NPU dataset is created with more presentation attack instruments (2 printers and 2 display devices vs. less in other datasets). (2) Oulu-NPU videos are recorded at Full HD resolution, i.e, $1920\times1080$, which is the resolution of high-quality videos of CASIA-MFSD. Therefore, resizing high-resolution images like Oulu-NPU will significantly blur the image and cause the performance drop.

\subsection{Effect of multi-target feature translation}

We further conduct extensive experiment to illustrate the effect of model extension. In addition to SS2ST and SS2MT, we also conduct \textbf{S}ingle \textbf{S}ource to \textbf{B}lending \textbf{T}argets (\textbf{SS2BT}), which applies the same model as SS2ST, but the target domain is a mixture of multiple target domains. We report the average HTER and AUC score of the three target datasets and the results are shown in Table \ref{tab:result_compare_ours}. It could be discovered that the results of SS2MT are significantly superior to those of SS2BT, which shows the single target liveness encoder is not capable of extracting a robust liveness feature for multiple sub-domains in a mixed target given the large variation between sub-domains. Therefore, regarding multiple sub-domains as separate individual domains, model extension will achieve better performance.

We also observe that the average HTER and AUC scores of SS2MT are better than those of SS2ST. It implies that liveness feature of source domain in SS2MT contains more abundant information compared with SS2ST. In fact, SS2MT adapts liveness feature from source domain to multiple target domains, driving it to capture liveness information from more scenarios. In addition, SS2MT requires only one training cycle to obtain the results for multiple target domains. 

\begin{table*}[!t]
    \centering
    \renewcommand\tabcolsep{1.5pt}
    \resizebox{1.75\columnwidth}{16.2mm}{
    \begin{tabular}{ccccccccccc}
        \specialrule{.1em}{.05em}{.05em}
        \noalign{\vskip 1.5pt}
         \textbf{Method} &  \textbf{I $\rightarrow$ C} & \textbf{M $\rightarrow$ C} & \textbf{O $\rightarrow$ C} & \textbf{C $\rightarrow$ I} & \textbf{M $\rightarrow$ I} & \textbf{O $\rightarrow$ I} & \textbf{C $\rightarrow$ M} & \textbf{I $\rightarrow$ M} & \textbf{O $\rightarrow$ M} & \textbf{Avg.} \\
         \noalign{\vskip 1.5pt}
         \hline
         \noalign{\vskip 1.5pt}
         CDFTN-L w/o $\mathcal{L}_{cls^L}$&12.9&11.9&19.9&2.3&16.0&9.9& 15.6&10.0& 9.4&12.0\\
         CDFTN-L w/o $\mathcal{L}^{cyc}$&12.4&14.4&23.8&1.9& 4.9&6.2&11.1&15.2& 9.3&11.0\\
         CDFTN-L w/o $\mathcal{L}^{re}$&47.8&24.2&19.7&3.6&10.6&7.4&17.0&11.1&35.5&19.7\\
         CDFTN-L w/o $\mathcal{L}^{lat}$&19.3&14.6&21.4&3.8&1.4&6.2&8.3&19.1&10.8& 11.7\\
         \noalign{\vskip 1.5pt}
         \specialrule{.1em}{.05em}{.05em}
         \textbf{CDFTN-L}                         &\textbf{11.9}&\textbf{8.8}&\textbf{19.1}&\textbf{1.7}&\textbf{1.3}&\textbf{5.8}&\textbf{8.1}&\textbf{9.6} &\textbf{6.3}&\textbf{8.1}\\
         \noalign{\vskip 1.5pt}
         \specialrule{.1em}{.05em}{.05em}
    \end{tabular}}
    \caption{Evaluation of each loss parts of our proposed framework on \textbf{C}, \textbf{I}, \textbf{O} and \textbf{M}.}
    \label{tab:ablation1}
\end{table*}

\subsection{Ablation Study} \label{sec:ablation}
We perform an ablation study to evaluate the contribution of each component of Equation (\ref{eq:stage1loss}) and compare the performance between two state-of-the-art binary classifiers: LGSC \cite{feng2020learning} and ResNet-18 \cite{he2016deep}. As also could be seen from the first five lines of Table \ref{tab:ablation1}, 1) we evaluate the performance of CDFTN w/o $\mathcal{L}_{cls^L}$ (CDFTN without optimization of Equation (\ref{eq:clsloss})), and the performance is worse than full model. This is because optimization of Equation (\ref{eq:clsloss}) drives $z^L_s$ to be discriminative. 2) CDFTN w/o $\mathcal{L}^{cyc}$ drops the optimization of cycle consistency loss, and its performance is worse than the full model. We consider that cycle consistency helps liveness feature transfer in a closed-loop instead of any random paths. 3). CDFTN w/o $\mathcal{L}^{re}$ removes the contribution made by reconstruction loss, also resulting in worse performances, which illustrates that self-reconstruction loss is also crucial to regularize the feature mapping process. 4). CDFTN w/o $\mathcal{L}^{lat}$ removes the contribution made by latent reconstruction loss, and the results are also worse than the full model; in fact, optimizing $\mathcal{L}^{lat}$ ensures that liveness features extracted from original input and pseudo-labeled images are identical. 

It could be discovered from Table \ref{tab:ablation1} that all loss components are crucial to achieving the optimal solution. Comparing all components, $\mathcal{L}^{re}$ is the most important as it aims at a robust reconstruction of original images during training; $\mathcal{L}^{lat}$ is the second most important component as it enforces the liveness encoders to extract unchanged liveness features when encoding original images and pseudo-labeled images.

\begin{figure}[!t]
    \centering
    \includegraphics[height=4.2cm]{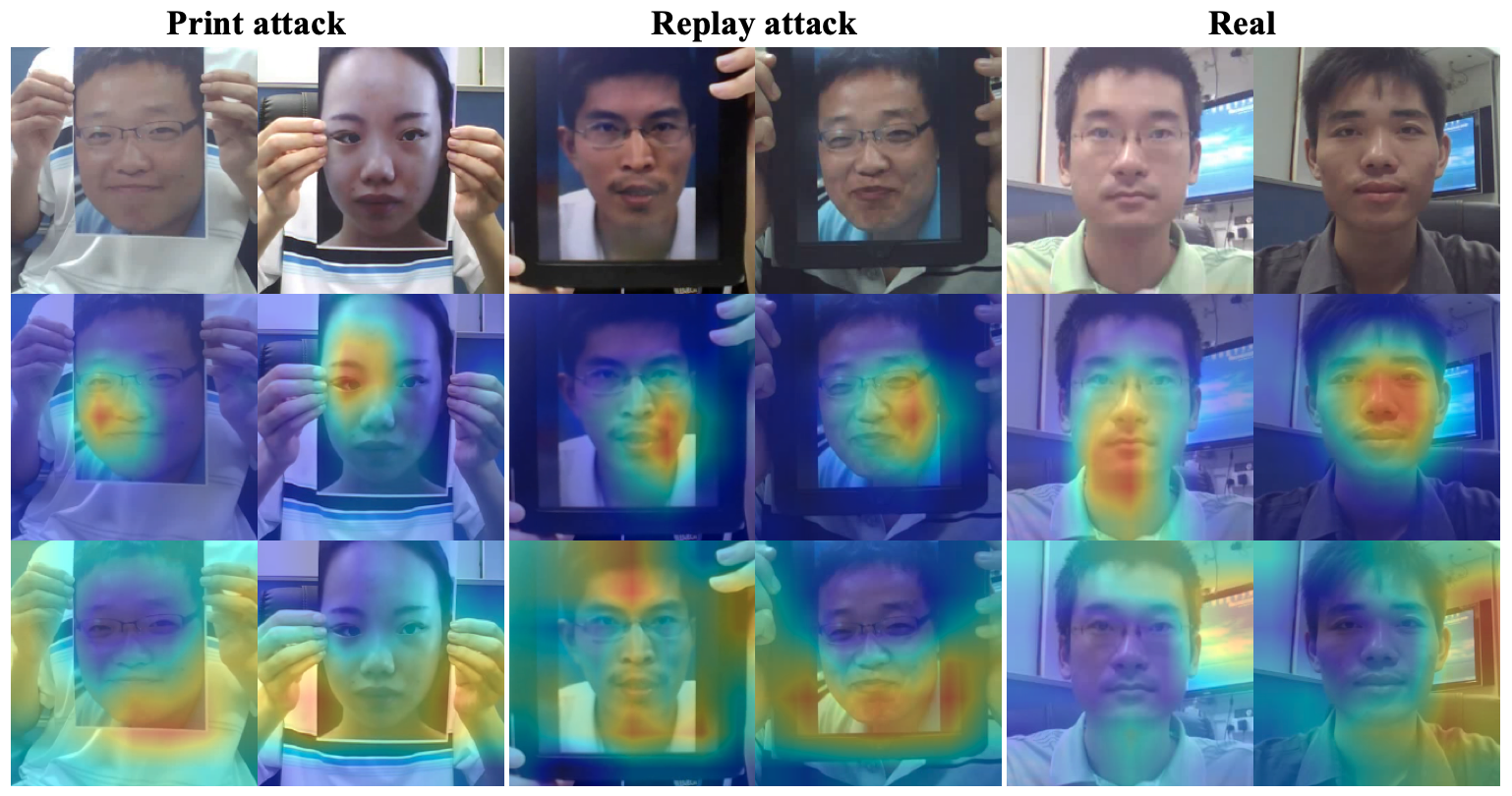}
    \caption{Grad-CAM \cite{Selvaraju2017Grad-CAM} visualizations of print attack, replay attack and real images by the ResNet-18 (middle row) and CDFTN-R method (bottom row) under the task $\textbf{I}\rightarrow\textbf{C}$. Original images are shown in the top row.}
    \label{fig:Grad-CAM}
\end{figure}

\begin{figure}[!t]
    \centering
    \includegraphics[height=4.2cm]{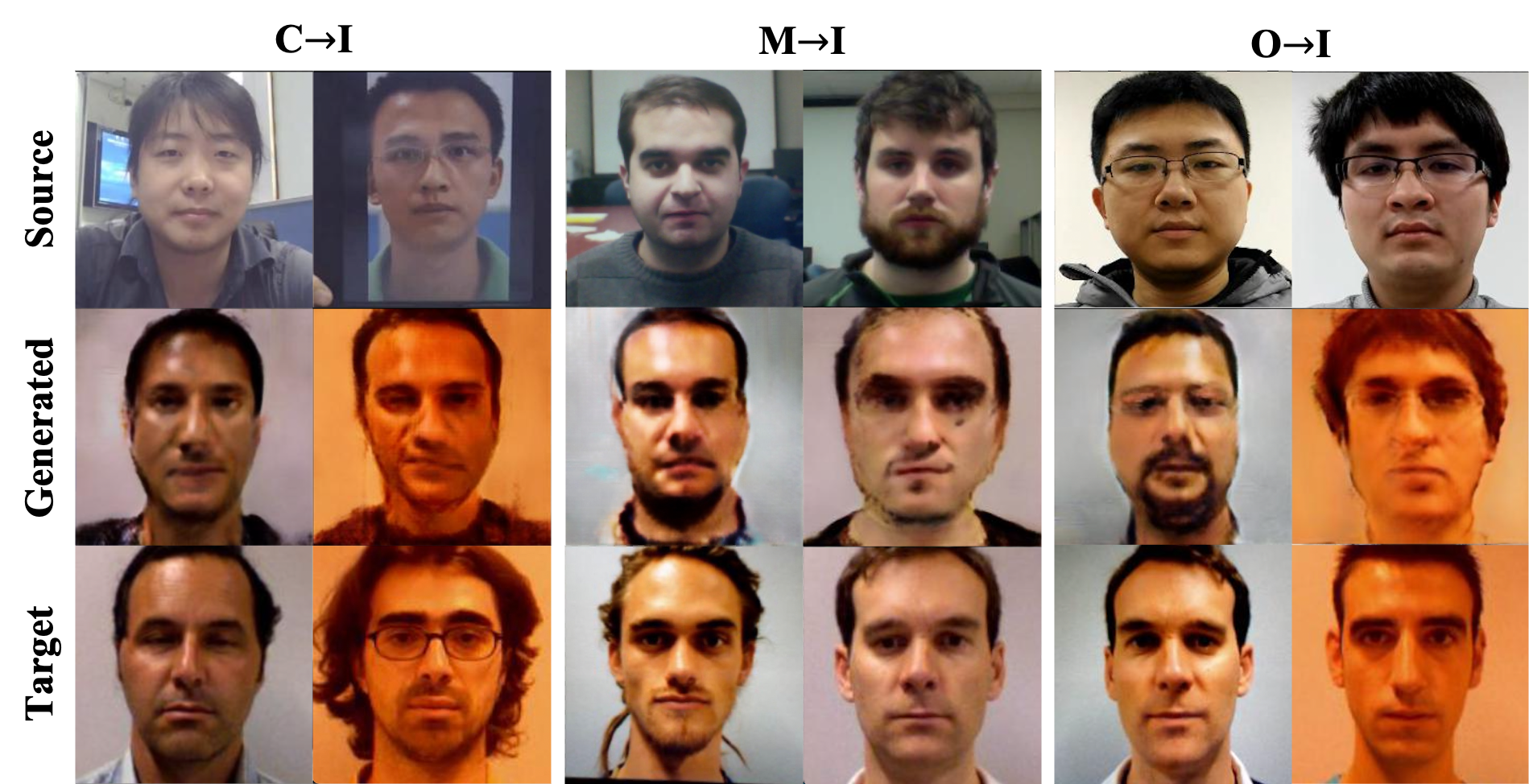}
    \caption{Visualizations of the generated images $\hat{x}_t$ (middle row) based on the liveness features of source domain  (top row) and content features of target domain (bottom row), i.e. $\hat{x}_t=G_t(z_s^L,z_t^C)$, given task $\textbf{C}\rightarrow\textbf{I}$, $\textbf{M}\rightarrow\textbf{I}$ and $\textbf{O}\rightarrow\textbf{I}$.}
    \label{fig:gen}
\end{figure}

\subsection{Visualizations}
Figure \ref{fig:gen} presents our generated images (middle row) based on liveness feature of source domain (top row) and content feature of target domain (bottom row). We train classifier $M$ on the generated images. From the perspective of image-to-image translation, faces in different domains possess different identities and backgrounds, thus do not fulfill a strict bijection relationship and cycle consistency cannot be completely satisfied. Therefore, the generated images appear to be a random mixture of both domains and do not have decent quality. However, our main purpose in this work is to improve the performance of cross-domain face anti-spoofing and the current generated images are good enough to possess cross-domain feature representations.

As shown in Figure \ref{fig:Grad-CAM}, we adopt Grad-CAM \cite{Selvaraju2017Grad-CAM} to show the class activation map (CAM) between the methods with and without CDFTN method. It shows that the classifier based on ResNet-18 mostly focuses on the face region. However, our method CDFTN-R pays more attention to the region of hands and the edge of paper or screen.

\begin{figure}
    \centering
    \begin{subfigure}{.495\linewidth}
        \centering
        \includegraphics[width=\linewidth]{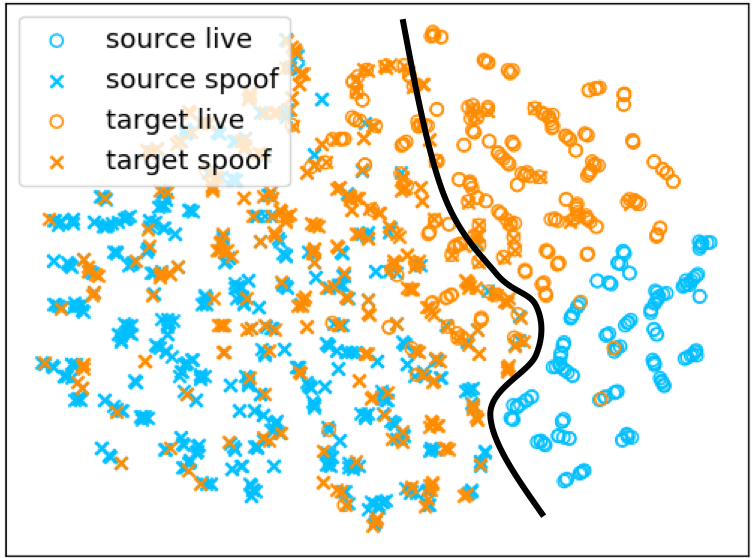}
        \caption{Before translation}
        \label{fig:tsne_scan_m-i}
    \end{subfigure}
    \begin{subfigure}{.495\linewidth}
        \centering
        \includegraphics[width=\linewidth]{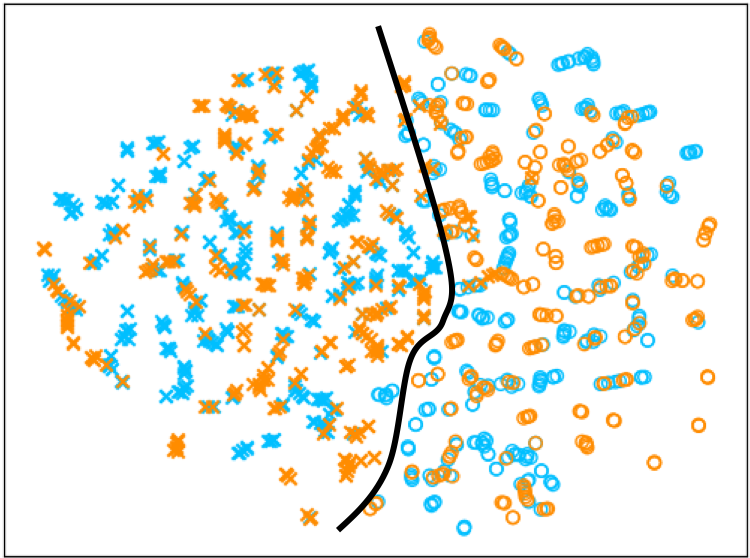}
        \caption{After translation}
        \label{fig:tsne_cdftn_m_i}
    \end{subfigure}
    \caption{The t-SNE \cite{maaten2008visualizing} visualization of feature distributions before and after feature translation upon task $\textbf{M}\rightarrow\textbf{I}$. Blue and orange points represent source and target domain, respectively. Circle and cross points represent live and spoof samples, respectively. }
    \label{fig:tsne}
\end{figure}

\subsection{Qualitative Analysis}
Considering the task of M $\rightarrow$ I, we visualize the feature distribution learned before and after feature translation to evaluate the optimization of domain divergence in Figure \ref{fig:tsne}(a) and Figure \ref{fig:tsne}(b), respectively. We randomly select 500 instances from each domains and plot t-SNE \cite{maaten2008visualizing} graphs. Comparing Figure \ref{fig:tsne}(a) and Figure \ref{fig:tsne}(b), we can discover that after adaptation, domain components are merged better than before, showing similar distributions between source and target domains. In addition, from the perspective of discriminative capability, a much clearer decision boundary is presented after feature translation.

\section{Conclusion} \label{sec:conclusion}
Cross-scenario face anti-spoofing remains a challenging task due to large variations in domain-specific features. In this work, we propose CDFTN to improve the current DA methods on FAS tasks. CDFTN achieves feature translation through swapping domain-invariant liveness features within domains. To achieve cyclic reconstruction, we propose to apply cycle consistency, self-reconstruction, and latent reconstruction modules. We train a classifier on pseudo-labeled images that generalizes well to target domain. Our experiments focus on evaluating the cross-database FAS performance and verify that our proposed method outperforms the state-of-the-art methods on various public datasets.

\bibliography{cdftn-aaai23}

\end{document}